%% file: main.tex
\documentclass[12pt]{article}

\usepackage{style}
\usepackage{commands}
\begin{document}

\begin{titlepage}
    \begin{center}
    
        \bigskip

\title{
  \vspace{-1cm}
  \hrule height 4pt
  \vskip 0.5cm
  \textbf{Calibration of Neural Networks}
  \vskip 0.5cm
  \hrule height 1pt
  \vskip 0.5cm
}
        \date{}
        \author{
            \textbf{Ruslan Vasilev}\thanks{
            Work done in 2020--2021 at Lomonosov Moscow State University as Ruslan Vasilev coursework (3\textsuperscript{rd} year) supervised by Alexander D'yakonov.
            } \\
            \texttt{artnitolog@yandex.com}
            \and
            \textbf{Alexander D'yakonov}\footnotemark[1] \\
            \texttt{djakonov@mail.ru}
        }

        {\let\newpage\relax\maketitle}
        \thispagestyle{empty}

    
        \begin{abstract}
            Neural networks solving real-world problems are often required not only to make accurate predictions but also to provide a confidence level in the forecast. The calibration of a model indicates how close the estimated confidence is to the true probability. This paper presents a survey of confidence calibration problems in the context of neural networks and provides an empirical comparison of calibration methods. We analyze problem statement, calibration definitions, and different approaches to evaluation: visualizations and scalar measures that estimate whether the model is well-calibrated. We review modern calibration techniques: based on post-processing or requiring changes in training. Empirical experiments cover various datasets and models, comparing calibration methods according to different criteria.
        \end{abstract}
        
    \end{center}

\end{titlepage}

{
    \hypersetup{linkcolor=black}
    \tableofcontents
}

\newpage

\section{Introduction}


Deep learning is finding more and more applications in various fields. Neural networks are actively used in clinical practice \cite{medical_nn}, self-driving cars \cite{self_driving}, machine translation \cite{nmt_google}, and other diverse applications.


Oftentimes, real-world problems require models that produce not only correct prediction but also a reliable measure of confidence in it. \emph{Confidence} refers to probability estimate that the forecast is correct. For example, if an algorithm predicts a given sample of patients are healthy with confidence 0.9, we expect that $90\%$ of them are really healthy. A model with reliable confidence estimation is called \emph{calibrated}. Along with the interpretation of neural network predictions, confidence calibration is important when probability estimates are fed into subsequent algorithm steps (for example, language models decoding strategies \cite{decode_lm}).


Modern neural networks often turn out to be poorly calibrated \cite{on_cal}. However, many other machine learning algorithms also produce biased confidence estimates \cite{good_proba, emp_comparison}. Various calibration techniques have been proposed for ``classical'' machine learning, some of which have been developed in deep learning.

\section{Problem Statement}


Consider a classification problem for objects from set $\mathcal{X}$ with classes ${\mathcal{Y}}=\{1,\dots,K\}$. Suppose we have trained a \emph{model}~--- an algorithm which produces vector of scores \emph{(confidences)} $\vec{a}(x)=(a_1(x),\dots,a_K(x))$, $\sum_{j=1}^{K}a_j(x)=1$, for each $x\in{\mathcal{X}}$. Next, the class corresponding to the highest confidence is assigned to the object:

\begin{equation*}\hat{y}(x)\coloneqq\underset{j\in \mathcal{Y}}{\argmax{a_j}},\quad \hat{p}(x)\coloneqq a_{\hat{y}}.
\end{equation*}


We would like to interpret estimator $\hat{p}$ as the probability that the true label coincides with the predicted one. If the estimate is accurate enough, then the model is called \emph{calibrated}. Formally, the definition of calibration (\emph{perfect calibration} in \cite{on_cal}) can be written as follows:

\begin{equation}\label{eq:perfect_cal_guo}
    \mathbb{P}\left(y=\hat{y}\mid \hat{p}=p\right)=p \quad \forall p\in\left[0, 1\right].
\end{equation}


There are stronger definitions of model calibration\footnote{It should be noted that the term \emph{calibration} also often refers to \emph{methods} which make model confidences accurate.} than \eqref{eq:perfect_cal_guo}. For example, according to \cite{isotonic}, the classifier is called calibrated (originally, \emph{well-calibrated}) if

\begin{equation}\label{eq:cw_cal}
    \mathbb{P}(y=j\mid a_j=p) = p \quad \forall j\in\mathcal{Y}, \quad \forall p\in \left[0, 1\right]
\end{equation}
that means the assurances given for each class (not just the predicted one) are calibrated.


In the case of real-world data and models, we cannot directly check \eqref{eq:perfect_cal_guo} and \eqref{eq:cw_cal}, so various calibration metrics come in handy as well as visualizations that are reviewed in \autoref{sec:estimate}.


\hyperref[sec:methods]{Section 4} describes calibration methods~--- techniques that make confidences more reliable. First, one can calibrate confidences afterwards, i.e. find a transformation that maps biased estimates to calibrated. There are different algorithms that find optimal transformations. Second, it is possible to apply special techniques during training, for example, loss function modifications.


\hyperref[sec:experiments]{Section 5} provides empirical comparison of calibration methods for modern neural network architectures and shows how the choice of loss function affects the calibration.

\section{Calibration Evaluation}\label{sec:estimate}
\subsection{Visualization}


Before defining calibration measures, we simplify the problem to binary classification: $\mathcal{Y}=\{0,1\}$~--- consider the model generates confidences $\hat{p}$ that the object belongs to the positive class \textit{($y=1$)}. Binary classification is common in various applications and ``classical'' machine learning models: logistic regression, support vector machines, gradient boosted trees and others~--- the problem of their calibration were studied in \cite{good_proba, emp_comparison}.


\mpl{rel_intro}{Options for visualizing confidence calibration. For clarity, synthetic data was generated. Support vector machine is used as a model (the distances to the separating hyperplane are scaled into $[0, 1]$).}


Consider a set of objects \textit{(usually, another validation set)}, for each of which the true class is known and the confidence score is produced by the model. Divide the segment of all possible confidence values $[0, 1]$ into $M$ equal-width intervals $I_m$:

\begin{equation}\label{eq:binning}
I_1= \left[0, \frac{1}{M}\right),\
I_2= \left[\frac{1}{M},\frac{2}{M}\right),\
\dots,\
I_{M-1}= \left[\frac{M-2}{M},\frac{M-1}{M}\right),\
I_{M} = \left[\frac{M-1}{M}, 1\right].
\end{equation}


Thus, each confidence estimate falls into one of these intervals~--- let $B_m$ denote the set of indices of those objects which confidences are in $I_m$. Both $B_m$ and $I_m$ are called \emph{bins}.


For each bin $B_m$, we calculate positive frequency $A^1_m$ and average confidence $C^1_m$:

\begin{equation}\label{eq:bin_accconf}
    A^1_m=\frac{1}{|B_m|}\sum_{i\in B_m} \mathbb{1}(y_i=1),
    \quad
    C^1_m=\frac{1}{|B_m|}\sum_{i\in B_m} \hat{p}_i.
\end{equation}


Finally, we can draw a plot $(C^1_m, A^1_m)_{m=1}^M$ which is called \emph{reliability plot} \autoref{fig:rel_intro} (a). Also, the resulting curve is sometimes called calibration curve. The model is considered well-calibrated if its calibration curve is close to the diagonal \textit{(in \autoref{subsec:calmetrics} we describe scalar metrics of such closeness)}.


Likewise, $(C^1_m, A^1_m)_{m=1}^M$ can be depicted using a histogram, which is called \emph{reliability diagram}. In \autoref{fig:rel_intro} (b) the average confidence is shown in red, and the positive frequency is shown in blue. If the red bar is higher than the blue one, then the algorithm underestimates confidences~--- \emph{underconfidence}. The opposite case is called \emph{overconfidence}. For better interpretation, the fraction of objects (bin \emph{weight}) that fell into the bin can also be shown in the graph.


When the number of classes $n>2$, reliability diagrams are built differently. The most popular approach corresponds to the definition \eqref{eq:perfect_cal_guo}. For each bin $B_m$, we estimate ``accuracy'' $A_m$ and average confidence $C_m$:

\begin{equation}\label{eq:accconf}
    A_m=\frac{1}{|B_m|}\sum_{i\in B_m} \mathbb{1}(y_i=\hat{y}_i),
    \quad
    C_m=\frac{1}{|B_m|}\sum_{i\in B_m} \hat{p}_i.
\end{equation}


The difference between \eqref{eq:bin_accconf} and \eqref{eq:accconf} is that in multiclass case $\hat{y}_i$ and $\hat{p}_i$ correspond to the estimated class and its confidence, while in binary case statistics are calculated only for a positive class.


Note that $A_m$ and $C_m$ estimate the left and the right of \eqref{eq:perfect_cal_guo}, respectively. They can be depicted on reliability diagram. For two classes, this approach is illustrated in \autoref{fig:rel_intro} (c). Bins to the left of 0.5 turn out to be empty because a binary classification algorithm assigns an object to a class that has confidence $>0.5$.


One can also consider classwise reliability diagrams \cite{dirichlet}. To make it, each class should be separately assigned to the positive, while all the others should be treated like the negative, so $n$ reliability diagrams for the binary case can be built. Although the classwise approach is more informative \eqref{eq:cw_cal}, when the number of classes is large (for example, 1'000 or 22'000 in ImageNet \cite{imagenet}), it is usually impractical due to the interpretation difficulty. Therefore, reliability diagrams considering only the prediction confidence \eqref{eq:accconf} are more common.

\subsection{Calibration Metrics}\label{subsec:calmetrics}


In addition to visualizations, various metrics\footnote{Here, \emph{metric} refers to measure for the evaluation of algorithms} can be used to evaluate the calibration of the model. One of the most common is Expected Calibration Error (ECE) \cite{bayesian_ece}. It estimates the expectation of the absolute difference between confidence and associated accuracy:

\begin{equation}\label{eq:ece_exp}
    \mathbb{E}_{\hat{p}}\left|
\mathbb{P}\left(y=\hat{y}\mid \hat{p}\right)-\hat{p}\right|.
\end{equation}

\eqref{eq:ece_exp} can be approximated using the partition of the confidences into bins ($l$ is the total number of objects in the considered set):
\begin{align}\label{eq:ece}
\ece &=
\sum_{m=1}^{M}
\frac{|B_m|}{n}
\left| A_m - C_m \right| \\
&=
\sum_{m=1}^{M}
\frac{|B_m|}{n}\left|
\frac{1}{|B_m|}\sum_{i\in B_m} \mathbb{1}(y_i=\hat{y}_i)
-
\frac{1}{|B_m|}\sum_{i\in B_m} \hat{p}_i
\right| \nonumber\\
&=
\frac{1}{n}\,\sum_{m=1}^{M}\,
\left|\,
\sum_{i\in B_m} \mathbb{1}(y_i=\hat{y}_i)
-
\sum_{i\in B_m} \hat{p}_i
\,\right|\nonumber.
\end{align}


Comparing \eqref{eq:ece} and reliability diagrams for a multiclass problem, we notice that ECE is exactly equal to the weighted average of the gaps between red and blue bars \autoref{fig:rel_intro}.

There are other metrics based on a partition of confidences into bins, although used less often. For example, one can calculate the maximum gap between confidence and accuracy \cite{bayesian_ece}:
\begin{equation}
    \mce=\max_m \left|A_m - C_m\right|,
\end{equation}
or take into account the confidence not only for the predicted class but also for all the others (classwise ECE) \cite{dirichlet}:
\begin{equation}\label{eq:cwece}
    \cwece=\frac{1}{K}
\sum_{j=1}^K \sum_{m=1}^M
\frac{|B^j_m|}{n}|A^j_m - C^j_m|,
\end{equation}
where $B^j_m, A^j_m, C^j_m$ are, respectively, $m$-th bin, accuracy and confidence, if we consider $j$-th class positive, and collect all the others into negative. This metric corresponds to classwise reliability diagrams.


Instead of equal-width bins \eqref{eq:binning}, bins with an equal number of samples can be used --- sometimes confidence diagrams are built in this way. In \cite{ace}, it was proposed to use equal-frequency bins to count the metrics described above. Further, an equal-width scheme will be considered. Also, in addition to $l_1$-norm (i.e. averaging modules), we can use $l_2$ \cite{verified_uncertainty}.


The problem with binning metrics is the dependence on the number of bins. An alternative approach is to use proper scoring rules. We consider Negative Log-Likelihood (NLL):

\begin{equation}\label{eq:nll}
    \nll=-\frac{1}{l}\sum_{i=1}^n \log{a_{i, y_i}}
\end{equation}
where $y_i$ is the true class label, $a_{i, y_i}$ is a confidence of the algorithm in it, $n$ is a total number of objects.

Another proper scoring rule, which can be used to evaluate model calibration, is Brier Score:
\begin{equation}
    \bs = \frac{1}{n}\sum_{i=1}^n \sum_{j=1}^K
    \left(a_{ij} - \mathbb{1}(y_i = j)\right)^2,
\end{equation}
where $K$ is a number of classes.

\section{Calibration Methods}\label{sec:methods}


There are two main types of calibration techniques. First, model outputs can be post-processed. Special transformation, \emph{calibration map}, maps biased probability estimates to the calibrated ones. Second, calibration can be incorporated into the model training itself.

\subsection{Post-processing}

The transformation is usually found on a hold-out set \textit{(calibration set)} $(x_i,y_i)_{i=1}^{n}$. It can be the same dataset used for hyperparameter tuning, but not the training set because model outputs distribution on training data is not the same as on unseen data.


\mpl{calibs_binary}{Different calibration maps for binary classification (the same data and model as in \autoref{fig:rel_intro}).}

\subsubsection{Histogram Binning}

The method was originally introduced in \cite{hist_binning} to calibrate decision trees and naive Bayes classifiers. A calibration map in histogram binning is piecewise constant. Consider the binary case: the set of output confidence values is divided into bins $B_1,\dots,B_M$  \emph{(usually equal-width \eqref{eq:binning} or equal-frequency)}, and scores that fall into $B_m$ are replaced with the common $\theta_m$. To find $\theta_1,\dots,\theta_M$, the following optimization problem is solved:

\begin{equation}\label{eq:hist_binning}
\sum_{m=1}^{M}\sum_{i\in B_m}
\left(\theta_m - y_i\right)^2 \ \to \
\min_{\theta_1,\dots, \theta_M}.
\end{equation}


In such a statement the optimal $\theta_m$ equals the fraction of positive objects that fall into $B_m$. The calibration map is illustrated in \autoref{fig:calibs_binary} (a).


The method is generalized to the multiclass case using the strategy \emph{one-vs-rest}: each class is separately declared positive and $K$ piecewise constant functions are constructed. In the inference stage, a calibrated vector is additionally normalized.

\subsubsection{Isotonic Regression}


The method was proposed in \cite{isotonic}. For the binary case, the map is piecewise constant again, but both the number of the intervals $M$ and their boundaries are optimized. The constraint is that the map should be non-decreasing. Thus, the following problem is solved:

\begin{equation}\label{eq:isotonic}
\sum_{m=1}^{M}\sum_{i\in \tilde{B}_m}
\left(\theta_m - y_i\right)^2 \ \to \
\underset{\substack{
    M \\
    \theta_1 \leqslant \dots \leqslant \theta_M \\
    0=\alpha_0 \leqslant \alpha_1 \leqslant \dots \leqslant \alpha_{M-1} \leqslant \alpha_M = 1
}}{\min,}
\end{equation}
where
$\tilde{B}_1 =\{i: \alpha_0 \leqslant \hat{p}_i < \alpha_1\},
\dots,
\tilde{B}_{m} =\{i: \alpha_{m-1} \leqslant \hat{p}_i \leqslant \alpha_m\}$.
The shape of the function is illustrated in \autoref{fig:calibs_binary}.


Isotonic regression is generalized to the multiclass case in the same way as histogram binning.

\subsubsection{Generalizations of Platt Calibration}


Originally, the method was proposed in \cite{platt} for calibration of the support vector machines. As can be seen in the illustrations \autoref{fig:rel_intro}, \autoref{fig:calibs_binary}, if we rescale the distances $r(x)$ from the objects to the separating hyperplane into $[0, 1]$ and treat them as confidences in positive class, then the reliability plot will have the form of a sigmoid:

\begin{equation}
    \hat{p}(x) = \frac{1}{1+e^{-(\alpha \cdot r(x) + \beta)}}.
\end{equation}


Scale parameter $\alpha$ and location parameter $\beta$ are optimized on a calibration set using maximum likelihood estimation. In this method, the transformation is continuous and allows different generalizations to the multiclass case.


The last linear layer of a neural network for the object $x$ outputs the logit vector: $\vec{z} = (z_1,\dots,z_K)$. To estimate class probabilities, the vector is transformed with softmax $\softmax(\cdot)$:

\begin{equation*}
    \softmax\left(\vec{z}\right)=
    \frac{1}{\sum_{j=1}^{K} {\exp{\left(z_j\right)}}}
    \left(
        {\exp{\left(z_1\right)}},
        \dots,
        {\exp{\left(z_K\right)}}
    \right),
\end{equation*}
so it is possible to generalize Platt Calibration by introducing scale and location parameters for logits:
\begin{equation}
    a(x) = \softmax(\vec{W}\cdot \vec{z} + \vec{b}).
\end{equation}


Parameters $\vec{W}$ and $\vec{b}$ are also optimized with maximum likelihood estimation on a calibration set, which is equivalent to minimizing NLL \eqref{eq:nll}. Depending on $\vec{W}$ and $\vec{b}$ shapes, different generalization may be defined:

\begin{enumerate}
    \item Temperature scaling:
    $$\vec{W}=\frac{1}{T}\in\mathbb{R}, \ T>0, \ \vec{b}=\vec{0}.$$
    

    Generalization of Platt Calibration with a single scalar parameter. The method is one of the most frequently used. An increase in temperature $T$ leads to an increase in uncertainty --- an increase in the entropy of the output distribution. Decreasing $T$, on the contrary, increases confidence in the predicted class. At the same time, the predicted class remains unchanged.

    \item Vector scaling:
    $$\vec{W}=\operatorname{diag}(\vec{v})\in \mathbb{R}^{K\times K}\text{ --- diagonal matrix}, \vec{v}\in\mathbb{R}^K.$$


    In this approach, a different scale factor is optimized for each class (and the bias, if $\vec{b}\neq \vec{0}$ is optimized too).

    \item Matrix scaling:
    $$\vec{W}\in \mathbb{R}^{K\times K}, \vec{b}\in\mathbb{R}^K.$$


    Matrix scaling is the most general parametrization in this group of methods and is equivalent to logistic regression in the logit space. However, with a large number of classes, the method has too many parameters, which can lead to overfitting, because a calibration set is usually not large.

\end{enumerate}

Note that to implement any of these methods, it is enough to add a linear layer (of the required dimension) to the frozen neural network.

\subsection{Calibration During Training}
The performance of neural networks strongly depends on the loss function. Usually, NLL \eqref{eq:nll} is used. Given an object $x$, it is equal to the cross-entropy between the true classification one-hot distribution $\vec{y}$ and the predicted distribution:
\begin{equation}
    \operatorname{CE}(\vec{y}, \vec{a}) = -\sum_{j=1}^{K} y_j \log{a_j}.
\end{equation}


To improve the calibration of the model, one can modify the loss function itself.

\subsubsection{Label Smoothing}


In this method, the degenerate distribution of the target is replaced by a smoothed one. The smoothing degree can be tuned using the parameter $\alpha\in [0, 1]$:

\begin{equation}
    \vec{y}=\left(y_1,\dots,y_K\right)
    \mapsto
    \left(
        (1-\alpha)y_1 + \frac{\alpha}{K},
        \dots,
        (1-\alpha)y_K + \frac{\alpha}{K}
    \right)=\vec{y}'.
\end{equation}


As parameter $\alpha$ increases, the distribution of $\vec{y}'$ becomes more uniform. After this transformation, $\operatorname{CE}(\vec{y}',\vec{a})$ the cross-entropy between the smoothed classification vector and the predicted distribution is minimized.


Although using of smoothed labels to train a classifier is not a new idea, this approach was proposed for calibration in \cite{smoothing}.

\subsubsection{Focal Loss}

\begin{figure}[!h]
    \includegraphics[width=0.7\textwidth]{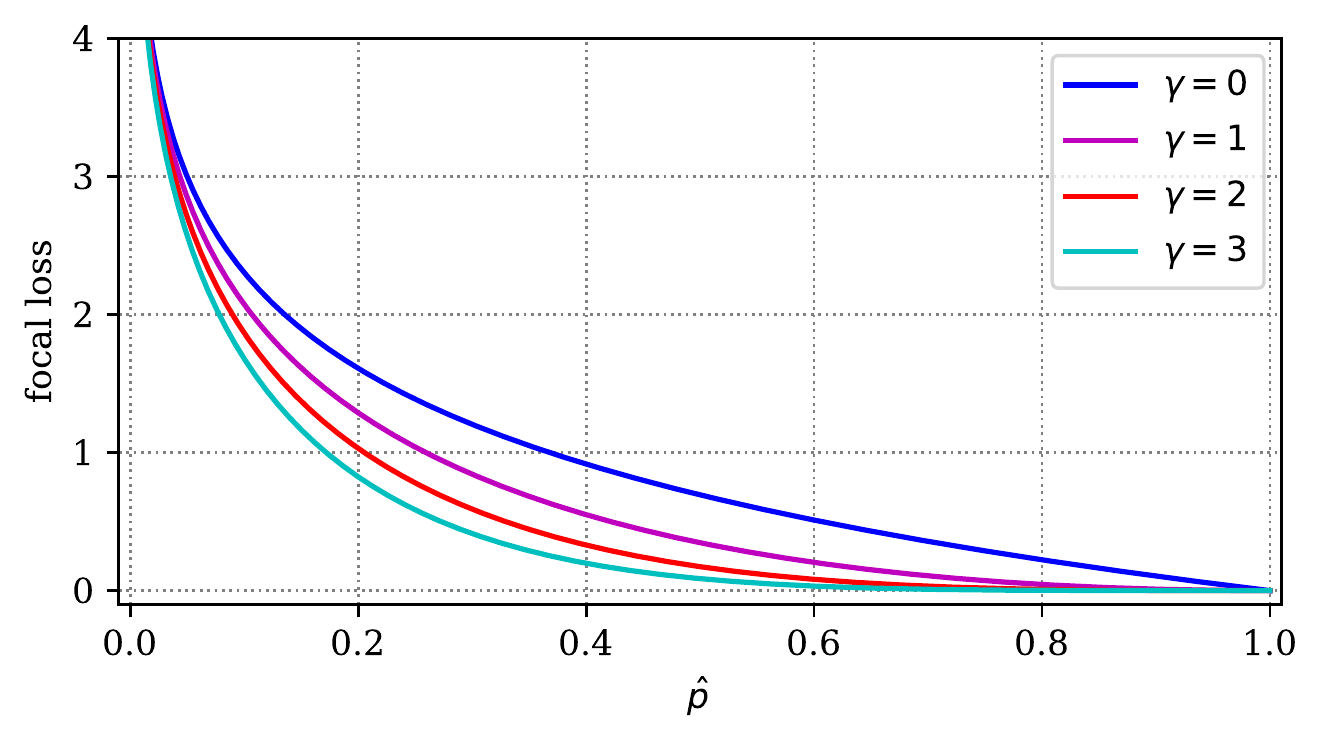}
    \centering
    \caption{Focal loss on a single object. $\hat{p}$ --- probability estimation for a true class}
    \label{fig:focal_loss}
\end{figure}


The focal loss was originally introduced to tackle the problem of class imbalance \cite{focal_detection}. In terms of confidence calibration, the idea was first used in \cite{focal_calib}. For an object belonging to the $j$-th class, the focal loss has the following definition:
\begin{equation}
    \operatorname{FL}=-(1-a_{j})^{\gamma}\cdot \log a_j, \quad \gamma \geqslant 0.
\end{equation}

Note that the loss function becomes the cross-entropy when $\gamma=0$. Increasing parameter $\gamma$, as we see in \autoref{fig:focal_loss}, decreases the penalty for those objects with already high confidence in the true class . While the cross-entropy is the upper bound of the Kullback--Leibler divergence between the true $\vec{y}$ and the predicted $\vec{a}$ distribution, the focal error has the entropy of the predicted distribution $H(\vec{a})$\cite{focal_calib} subtracted from the estimate:

\begin{equation*}
    \operatorname{CE}(\vec{y},\vec{a})
    \geqslant
    \operatorname{KL}(\vec{y}||\vec{a}),
    \qquad
    \operatorname{FL}(\vec{y},\vec{a})
    \geqslant
    \operatorname{KL}(\vec{y}||\vec{a})-\gamma\cdot \operatorname{H}(\vec{a}).
\end{equation*}


Thus, optimizing the focal error additionally increases the entropy of the predicted distribution and helps to tackle overconfidence.

\section{Empirical Experiments}\label{sec:experiments}

\subsection{Experimental Design}


The following datasets are used in the experiments:

\begin{itemize}

    \item \textbf{CIFAR-10} \cite{cifar}: The dataset consists of $60\,000$ color images $32\times 32$. \emph{Training} / \emph{validation} / \emph{test} splits are respectively $50\,000\;/\;5\,000\;/\;5\,000$.


    \item \textbf{CIFAR-100} \cite{cifar}: $60\,000$ color images $32\times 32$, 100 classes. \emph{Training} / \emph{validation} / \emph{test}: $50\,000\;/\;5\,000\;/\;5\,000$.


    \item \textbf{ImageNet 2012} \cite{imagenet}: Large dataset with color images organized into 1000 classes. \emph{Training} / \emph{validation} / \emph{test}: $1.2\;\text{m}\;/\;25\,000\;/\;25\,000$.


    \item \textbf{Tiny ImageNet} \cite{imagenet}: $110\,000$ color images $64\times 64$, organized into 200 classes, subset of the previous dataset. \emph{Training} / \emph{validation} / \emph{test}: $100\,000\;/\;5\,000\;/\;5\,000$.
\end{itemize}


Pre-trained neural networks with various architectures from open repositories were used for calculations. In experiments, models and datasets are divided into two main groups:

\begin{enumerate}

    \item The first group includes neural networks trained on CIFAR-10, CIFAR-100, and ImageNet. The weights for the models are obtained respectively from the repositories \cite{pretrained_cifar10, pretrained_cifar100, pretrained_imagenet}. Models from this group are used to compare calibration methods based on post-processing.


    \item The second group includes pre-trained neural networks from the repository \cite{focal_github}. The datasets used here are CIFAR-10, CIFAR-100, and Tiny ImageNet. These neural networks were trained in \cite{focal_calib}~--- focal error and label smoothing were used for the considered models.
\end{enumerate}


The models were trained on the \emph{training} data \textit{(or its subset)}, calibrated on the \emph{validation} set~--- all diagrams and metrics correspond to the \emph{test} set.


Code used in all experiments is published in \cite{my_repo}: temperature, vector, and matrix scaling are implemented using PyTorch, and other methods and evaluation are implemented with SciPY and scikit-learn. Histogram binning use 20 bins; ECE, cwECE, and MCE are calculated on 15 bins partition; reliability diagrams use 10 bins.

\subsection{Experiment Results}


Complete tables with measurements are given in 
\hyperref[sec:appendix]{appendix}, additional reliability diagrams for all the models considered in this work can be found in the accompanying repository \cite{my_repo}.

\begin{figure}[!h]
    \includegraphics[width=\textwidth]{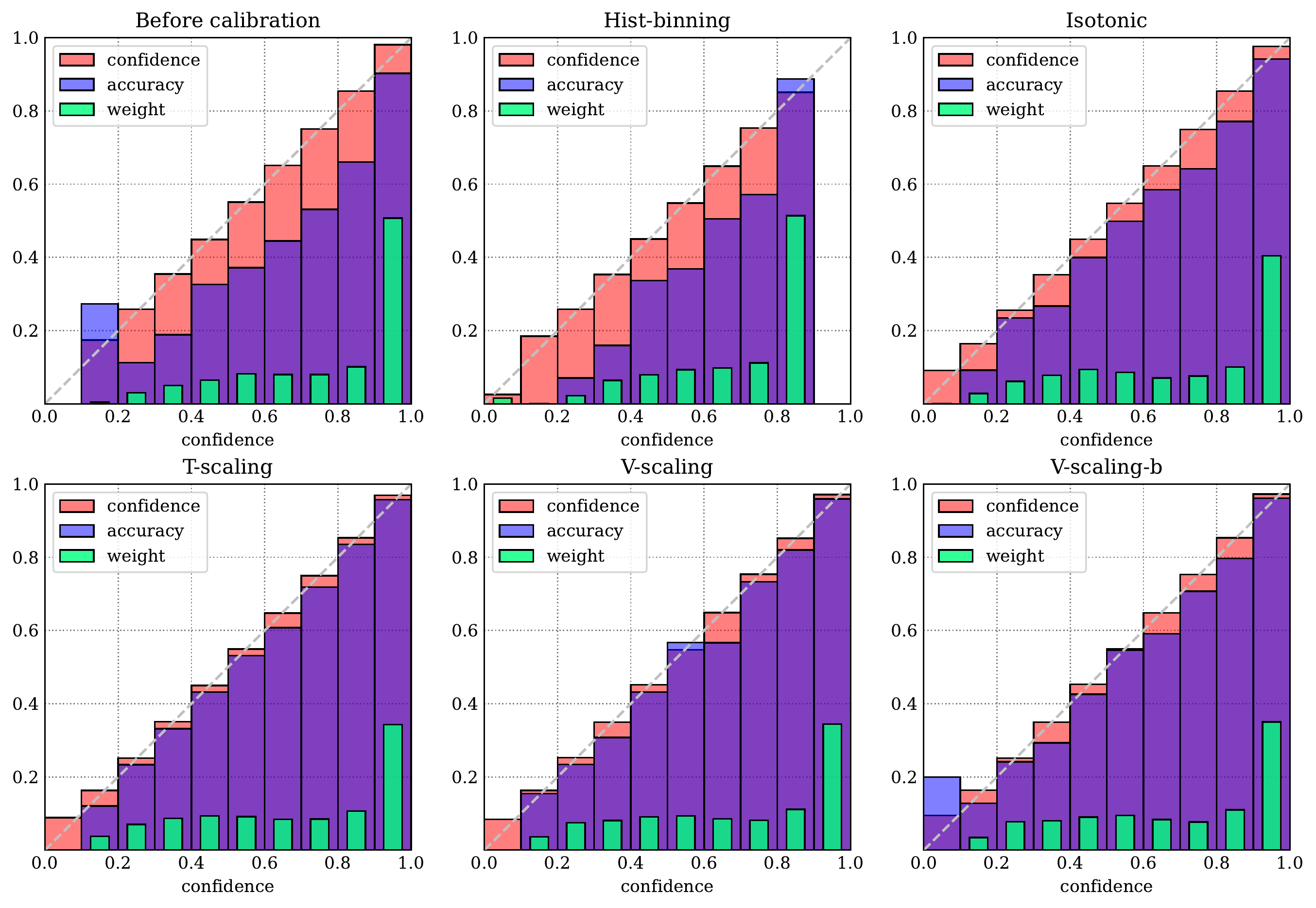}
    \centering
    \caption{CIFAR-100, ShuffleNetV2\_x0\_5}
    \label{fig:reldiag_1_21}
\end{figure}


Consider the reliability diagrams for ShuffleNetV2 (CIFAR-100, \autoref{fig:reldiag_1_21}): we see a ``typical'' state of modern neural networks calibration~--- overconfidence. Calibration methods help to correct the situation: in this case, temperature scaling works best for all the metrics. However, histogram binning changes probabilities too aggressively when the number of classes is large, as can be seen from the weights change. 

\begin{figure}[!h]
    \includegraphics[width=\textwidth]{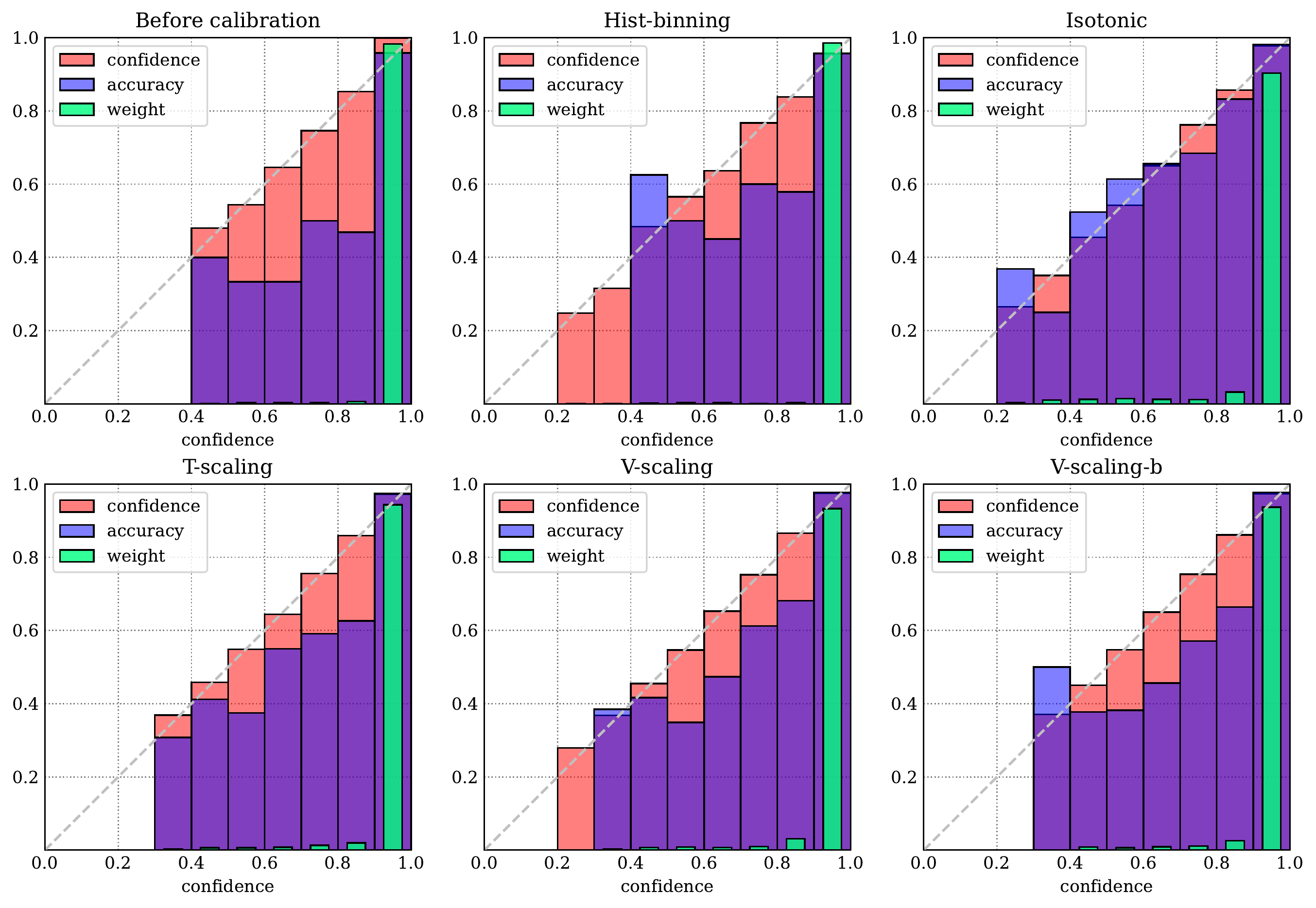}
    \centering
    \caption{CIFAR-10, DenseNet121 (CE)}
    \label{fig:reldiag_2_1}
\end{figure}

For a small number of classes, on the contrary, histogram binning works best in terms of confidence in prediction (\autoref{tab:metrics:ECE_1}, \autoref{tab:metrics:ECE_2}~--- for almost all models on CIFAR-10). Note that here neural networks already solve the classification problem with very high accuracy. Almost all probabilities of the predicted class are close to 1, as, for example, on \autoref{fig:reldiag_2_1}. In terms of MCE (\autoref{tab:metrics:MCE_1}, \autoref{tab:metrics:MCE_2})~--- metric, that doesn't consider weights of bins~--- histogram binning leads to low calibration.

For matrix scaling, we do not provide reliability diagrams: the method overfits too much when the number of classes is large. As a result, matrix scaling significantly degrades the classification quality (\autoref{tab:metrics:ACC_1}, \autoref{tab:metrics:ACC_2}) for all datasets, except, again, the low-class CIFAR-10.


One of the most common methods of neural network calibration is temperature scaling. The method doesn't affect classification predictions, while other calibration options almost always reduce accuracy (\autoref{tab:metrics:ACC_1}, \autoref{tab:metrics:ACC_2}).


In terms of NLL, temperature and vector scaling are the best as expected, because this loss was minimized during calibration: \autoref{tab:metrics:NLL_1}, \autoref{tab:metrics:NLL_2}. As for Brier Score, the best calibration method in many cases was isotonic regression: \autoref{tab:metrics:BS_1}, \autoref{tab:metrics:BS_2}.

\begin{table}[h!]
    \begin{minipage}[h!]{0.47\textwidth}
        \input{tabs/fl_ECE.tex}
    \end{minipage}\hfill
    \begin{minipage}[h!]{0.47\textwidth}
        \input{tabs/fl_cwECE.tex}
    \end{minipage}
\end{table}

\begin{figure}[!h]
    \includegraphics[width=\textwidth]{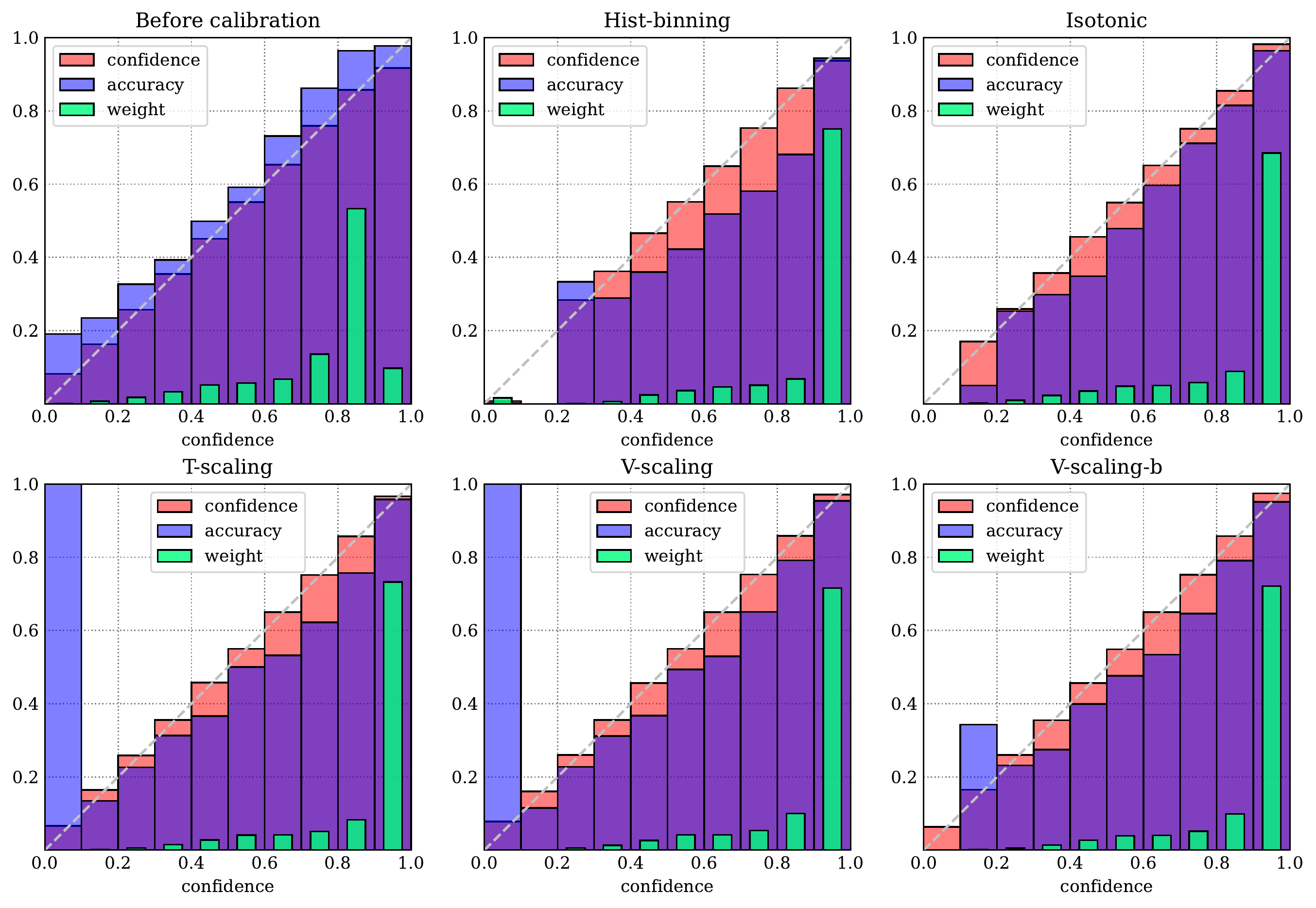}
    \centering
    \caption{ImageNet, EfficientNet\_b8}
    \label{fig:reldiag_1_29}
\end{figure}


Focal loss and label smoothing result in better calibrated models than the ones optimizing the cross-entropy~--- both in terms of calibration in the predicted class (\autoref{tab:fl:ECE}) and classwise estimates (\autoref{tab:fl:cwECE}). At the same time, such models can be further calibrated with post-processing. In the original paper \cite{focal_calib}, temperature scaling was used to calibrate models trained with focal loss. Although this approach does show good results with respect to ECE (\autoref{tab:metrics:ECE_2}), vector scaling works better for classwise cwECE (\autoref{tab:metrics:cwECE_2}). As for models of the first group (\autoref{tab:metrics:cwECE_1}), vector scaling also basically minimizes cwECE. These results are quite expected since vector scaling finds separate coefficients for each class.


Consider also reliability diagrams for EfficientNet (\autoref{fig:reldiag_1_29}). Among all the models used, the underconfidence is most clearly visible in this neural network: most of the predictions fall not into $[0.9, 1]$, but into $[0.8, 0.9)$. The reason for this behavior may be training setup: the model was trained with label smoothing ($\alpha=0.1$) \cite{pretrained_imagenet}. All calibration methods resulted in a noticeable increase in the confidence of the answers.

\section{Conclusion}


In this work, the main methods of confidence calibration were compared on different neural network architectures, datasets, and critera.


The applicability of a particular method depends significantly on the amount of data and the selected evaluation criterion. Algorithms, in which separate calibration maps are found for each class, work well only if there is a sufficient amount of data in the calibration set (this can usually be provided when the number of classes is small). Strategies based on linear transformation of logits (for example, temperature scaling) show high quality in problems with a large number of classes but are subject to overfitting when parameterization is excessive (matrix scaling).


Confidence calibration is still open to further research in machine learning: as shown in this work, not only calibration methods but also criteria choice can lead to different results.

\newpage
\printbibliography[
    heading=bibintoc,
    title={References}
]

\newpage
\begin{appendices}\label{sec:appendix}
\section{Classification Accuracy}
\input{tabs/metrics_ACC_1.tex}
\input{tabs/metrics_ACC_2.tex}
\clearpage

\section{Binning-based Metrics}
\input{tabs/metrics_ECE_1.tex}
\input{tabs/metrics_ECE_2.tex}
\input{tabs/metrics_cwECE_1.tex}
\input{tabs/metrics_cwECE_2.tex}
\input{tabs/metrics_MCE_1.tex}
\input{tabs/metrics_MCE_2.tex}
\clearpage

\section{Proper Scoring Rules}
\input{tabs/metrics_NLL_1.tex}
\input{tabs/metrics_NLL_2.tex}
\input{tabs/metrics_BS_1.tex}
\input{tabs/metrics_BS_2.tex}

\end{appendices}

\end{document}

%% file: tabs/fl_ECE.tex
\centering
\resizebox{\textwidth}{!}{\begin{tabular}{lllllll}
\toprule
     Dataset &            Model &    CE &  FL 1 &                  FL 2 &                  FL 3 &               LS 0.05 \\
\midrule
    CIFAR-10 &       DenseNet121 &  4.53 &  3.47 &                  2.02 &                  1.68 & \textbf{1.65} \\
    CIFAR-10 &         ResNet110 &  4.73 &  3.70 &                  2.78 & \textbf{1.61} &                  2.20 \\
    CIFAR-10 &          ResNet50 &  4.26 &  3.88 &                  2.55 & \textbf{1.58} &                  3.07 \\
    CIFAR-10 & Wide-ResNet-26-10 &  3.25 &  2.66 & \textbf{1.57} &                  1.98 &                  4.33 \\
   CIFAR-100 &       DenseNet121 & 20.90 & 14.54 &                  8.40 & \textbf{4.49} &                 13.27 \\
   CIFAR-100 &         ResNet110 & 19.76 & 15.35 &                 12.10 & \textbf{9.22} &                 11.44 \\
   CIFAR-100 &          ResNet50 & 18.14 & 13.36 &                  8.60 & \textbf{4.99} &                  8.15 \\
   CIFAR-100 & Wide-ResNet-26-10 & 16.28 &  9.12 &                  4.22 & \textbf{2.20} &                  5.27 \\
TinyImageNet &          ResNet50 & 15.98 &  7.87 &                  3.32 & \textbf{1.93} &                 15.73 \\
\bottomrule
\end{tabular}%
}
\caption{ECE, \% -- Expected Calibration Error Error (lower is better), 15 bins without post-processing, columns correspond to different loss functions}
\label{tab:fl:ECE}

%% file: tabs/fl_cwECE.tex
\centering
\resizebox{\textwidth}{!}{\begin{tabular}{lllllll}
\toprule
     Dataset &            Model &    CE &  FL 1 &                   FL 2 &                   FL 3 & LS 0.05 \\
\midrule
    CIFAR-10 &       DenseNet121 & 0.948 & 0.755 & \textbf{0.514} &                  0.524 &   0.576 \\
    CIFAR-10 &         ResNet110 & 0.990 & 0.804 &                  0.660 & \textbf{0.505} &   0.673 \\
    CIFAR-10 &          ResNet50 & 0.941 & 0.836 &                  0.625 & \textbf{0.524} &   0.766 \\
    CIFAR-10 & Wide-ResNet-26-10 & 0.699 & 0.611 & \textbf{0.479} &                  0.523 &   0.869 \\
   CIFAR-100 &       DenseNet121 & 0.458 & 0.364 &                  0.280 & \textbf{0.254} &   0.315 \\
   CIFAR-100 &         ResNet110 & 0.433 & 0.372 &                  0.321 & \textbf{0.281} &   0.299 \\
   CIFAR-100 &          ResNet50 & 0.412 & 0.337 &                  0.282 & \textbf{0.256} &   0.271 \\
   CIFAR-100 & Wide-ResNet-26-10 & 0.372 & 0.264 & \textbf{0.218} &                  0.226 &   0.239 \\
TinyImageNet &          ResNet50 & 0.250 & 0.218 &                  0.205 & \textbf{0.203} &   0.231 \\
\bottomrule
\end{tabular}%
}
\caption{cwECE, \% -- Classwise Expected Calibration Error (lower is better), 15 bins without post-processing, columns correspond to different loss functions}
\label{tab:fl:cwECE}

%% file: tabs/metrics_ACC_1.tex
\begin{table}[h!]
\centering
\resizebox{\textwidth}{!}{
\begin{tabular}{llccccccc}
\toprule
  Dataset &            Model &          Before calibration &           Hist-binning &               Isotonic &              T-scaling &              V-scaling &            V-scaling-b &            M-scaling-b \\
\midrule
 CIFAR-10 &       DenseNet121 & \textbf{93.92} &                  93.52 &                  93.72 & \textbf{93.92} &                  93.84 &                  93.80 &                  93.80 \\
 CIFAR-10 &       DenseNet161 &                  93.70 & \textbf{93.74} &                  93.58 &                  93.70 &                  93.72 & \textbf{93.74} &                  93.52 \\
 CIFAR-10 &       DenseNet169 & \textbf{94.08} &                  93.44 &                  93.84 & \textbf{94.08} &                  94.02 &                  93.94 &                  93.78 \\
 CIFAR-10 &         GoogleNet & \textbf{92.92} &                  92.58 &                  92.68 & \textbf{92.92} &                  92.90 &                  92.84 &                  92.76 \\
 CIFAR-10 &       InceptionV3 & \textbf{93.32} &                  93.12 &                  93.28 & \textbf{93.32} &                  93.22 & \textbf{93.32} &                  93.26 \\
 CIFAR-10 &       MobileNetV2 & \textbf{93.42} &                  93.40 & \textbf{93.42} & \textbf{93.42} &                  93.26 &                  93.36 &                  93.36 \\
 CIFAR-10 &          ResNet18 & \textbf{92.54} &                  92.14 &                  92.34 & \textbf{92.54} &                  92.48 &                  92.40 &                  92.18 \\
\bottomrule
\end{tabular}%
}
\caption{Accuracy \% (higher is better) -- fraction of correct predictions, group 1}
\label{tab:metrics:ACC_1}
\end{table}

%% file: tabs/metrics_ACC_2.tex
\begin{table}[h!]
\centering
\resizebox{\textwidth}{!}{\begin{tabular}{llccccccc}
\toprule
  Dataset &            Model &          Before calibration &           Hist-binning &               Isotonic &              T-scaling &              V-scaling &            V-scaling-b &            M-scaling-b \\
\midrule
    CIFAR-10 &            DenseNet121 (CE) &                  94.92 & \textbf{94.98} &                  94.84 &                  94.92 &                  94.94 &                  94.96 &                  94.76 \\
    CIFAR-10 &          DenseNet121 (FL 1) &                  94.84 &                  94.68 &                  94.82 &                  94.84 &                  95.00 & \textbf{95.02} &                  94.52 \\
    CIFAR-10 &          DenseNet121 (FL 2) &                  94.90 &                  94.60 & \textbf{94.96} &                  94.90 &                  94.84 &                  94.80 &                  94.80 \\
    CIFAR-10 &          DenseNet121 (FL 3) &                  94.24 &                  94.24 &                  94.20 &                  94.24 &                  94.34 &                  94.24 & \textbf{94.36} \\
    CIFAR-10 &       DenseNet121 (LS 0.05) &                  94.52 &                  94.62 & \textbf{94.66} &                  94.52 &                  94.56 &                  94.58 &                  94.62 \\
    CIFAR-10 &              ResNet110 (CE) &                  94.82 &                  94.70 &                  94.76 &                  94.82 &                  94.84 &                  94.86 & \textbf{94.90} \\
    CIFAR-10 &            ResNet110 (FL 1) &                  94.80 &                  94.90 &                  94.84 &                  94.80 &                  94.86 & \textbf{95.00} &                  94.98 \\
\bottomrule
\end{tabular}%
}
\caption{Accuracy \% (higher is better) -- fraction of correct predictions, group 2}
\label{tab:metrics:ACC_2}
\end{table}

%% file: tabs/metrics_ECE_1.tex
\begin{table}[h!]
\centering
\resizebox{\textwidth}{!}{\begin{tabular}{llccccccc}
\toprule
Dataset &            Model &          Before calibration &           Hist-binning &               Isotonic &              T-scaling &              V-scaling &            V-scaling-b &            M-scaling-b \\
\midrule
 CIFAR-10 &       DenseNet121 &                  1.86 & \textbf{1.08} &                  2.10 &                  1.64 &                  1.67 &                  1.58 &        1.49 \\
 CIFAR-10 &       DenseNet161 &                  2.11 & \textbf{1.08} &                  1.64 &                  1.68 &                  1.87 &                  1.75 &        1.75 \\
 CIFAR-10 &       DenseNet169 &                  2.04 & \textbf{1.42} &                  1.87 &                  1.72 &                  1.64 &                  1.62 &        1.64 \\
 CIFAR-10 &         GoogleNet &                  1.65 & \textbf{0.89} &                  1.36 &                  1.21 &                  1.46 &                  1.37 &        1.34 \\
 CIFAR-10 &       InceptionV3 &                  2.06 & \textbf{1.35} &                  1.91 &                  1.88 &                  1.99 &                  1.91 &        1.86 \\
 CIFAR-10 &       MobileNetV2 &                  2.92 & \textbf{1.45} &                  1.74 &                  1.87 &                  2.09 &                  1.95 &        1.78 \\
 CIFAR-10 &          ResNet18 &                  2.51 & \textbf{1.60} &                  2.03 &                  2.13 &                  1.98 &                  1.93 &        2.16 \\
\bottomrule
\end{tabular}%
}
\caption{ECE \% -- Expected Calibration Error (lower is better), 15 bins, group 1}
\label{tab:metrics:ECE_1}
\end{table}

%% file: tabs/metrics_ECE_2.tex
\begin{table}[h!]
\centering
\resizebox{\textwidth}{!}{\begin{tabular}{llccccccc}
\toprule
Dataset &            Model &          Before calibration &           Hist-binning &               Isotonic &              T-scaling &              V-scaling &            V-scaling-b &            M-scaling-b \\
\midrule
    CIFAR-10 &            DenseNet121 (CE) &                  4.53 & \textbf{0.38} &                  1.24 &                  1.64 &                  1.19 &                  1.29 &                  1.45 \\
    CIFAR-10 &          DenseNet121 (FL 1) &                  3.47 &                  0.87 & \textbf{0.65} &                  1.26 &                  0.88 &                  1.05 &                  1.44 \\
    CIFAR-10 &          DenseNet121 (FL 2) &                  2.02 &                  0.97 & \textbf{0.73} &                  0.95 &                  1.08 &                  1.01 &                  1.18 \\
    CIFAR-10 &          DenseNet121 (FL 3) &                  1.68 & \textbf{1.17} &                  1.32 &                  1.49 &                  1.32 &                  1.38 &                  1.84 \\
    CIFAR-10 &       DenseNet121 (LS 0.05) &                  1.65 & \textbf{0.93} &                  1.55 &                  1.22 &                  1.28 &                  1.24 &                  1.29 \\
    CIFAR-10 &              ResNet110 (CE) &                  4.73 & \textbf{1.01} &                  1.11 &                  1.23 &                  1.36 &                  1.39 &                  1.25 \\
    CIFAR-10 &            ResNet110 (FL 1) &                  3.70 & \textbf{0.71} &                  0.99 &                  1.11 &                  1.20 &                  1.06 &                  1.05 \\
\bottomrule
\end{tabular}%
}
\caption{ECE \% -- Expected Calibration Error (lower is better), 15 bins, group 2}
\label{tab:metrics:ECE_2}
\end{table}

%% file: tabs/metrics_cwECE_1.tex
\begin{table}[h!]
\centering
\resizebox{\textwidth}{!}{\begin{tabular}{llccccccc}
\toprule
Dataset &            Model &          Before calibration &           Hist-binning &               Isotonic &              T-scaling &              V-scaling &            V-scaling-b &            M-scaling-b \\
\midrule
 CIFAR-10 &       DenseNet121 &         0.513 &                  0.509 &                  0.566 &                  0.514 &                  0.548 &                  0.491 & \textbf{0.483} \\
 CIFAR-10 &       DenseNet161 &         0.646 &                  0.536 &                  0.534 &                  0.629 &                  0.531 & \textbf{0.527} &                  0.536 \\
 CIFAR-10 &       DenseNet169 &         0.551 &                  0.566 &                  0.527 &                  0.536 & \textbf{0.485} &                  0.490 &                  0.518 \\
 CIFAR-10 &         GoogleNet &         0.641 &                  0.579 &                  0.543 &                  0.559 &                  0.544 & \textbf{0.520} &                  0.545 \\
 CIFAR-10 &       InceptionV3 &         0.596 &                  0.573 & \textbf{0.543} &                  0.598 &                  0.543 &                  0.573 &                  0.553 \\
 CIFAR-10 &       MobileNetV2 &         0.638 & \textbf{0.500} &                  0.502 &                  0.543 &                  0.572 &                  0.574 &                  0.546 \\
 CIFAR-10 &          ResNet18 &         0.649 &                  0.578 &                  0.570 &                  0.634 &                  0.590 & \textbf{0.538} &                  0.581 \\
\bottomrule
\end{tabular}%
}
\caption{cwECE \% -- Classwise Expected Calibration Error (lower is better), 15 bins, group 1}
\label{tab:metrics:cwECE_1}
\end{table}

%% file: tabs/metrics_cwECE_2.tex
\begin{table}[h!]
\centering
\resizebox{\textwidth}{!}{\begin{tabular}{llccccccc}
\toprule
Dataset &            Model &          Before calibration &           Hist-binning &               Isotonic &              T-scaling &              V-scaling &            V-scaling-b &            M-scaling-b \\
\midrule
    CIFAR-10 &            DenseNet121 (CE) &         0.948 &                  0.478 &                  0.530 &                  0.548 &                  0.485 & \textbf{0.455} &                  0.461 \\
    CIFAR-10 &          DenseNet121 (FL 1) &         0.755 &                  0.487 & \textbf{0.376} &                  0.462 &                  0.430 &                  0.416 &                  0.540 \\
    CIFAR-10 &          DenseNet121 (FL 2) &         0.514 &                  0.488 &                  0.424 &                  0.462 & \textbf{0.419} &                  0.431 &                  0.428 \\
    CIFAR-10 &          DenseNet121 (FL 3) &         0.524 &                  0.489 & \textbf{0.474} &                  0.532 &                  0.508 &                  0.483 &                  0.517 \\
    CIFAR-10 &       DenseNet121 (LS 0.05) &         0.576 &                  0.376 &                  0.446 &                  0.514 &                  0.410 & \textbf{0.317} &                  0.392 \\
    CIFAR-10 &              ResNet110 (CE) &         0.990 &                  0.491 &                  0.547 &                  0.550 &                  0.495 & \textbf{0.449} &                  0.476 \\
    CIFAR-10 &            ResNet110 (FL 1) &         0.804 &                  0.483 & \textbf{0.404} &                  0.517 &                  0.511 &                  0.449 &                  0.458 \\
\bottomrule
\end{tabular}%
}
\caption{cwECE \% -- Classwise Expected Calibration Error (lower is better), 15 bins, group 2}
\label{tab:metrics:cwECE_2}
\end{table}

%% file: tabs/metrics_MCE_1.tex
\begin{table}[h!]
\centering
\resizebox{\textwidth}{!}{\begin{tabular}{llccccccc}
\toprule
Dataset &            Model &          Before calibration &           Hist-binning &               Isotonic &              T-scaling &              V-scaling &            V-scaling-b &            M-scaling-b \\
\midrule
 CIFAR-10 &       DenseNet121 &         42.22 &        32.98 &                  80.26 &                  41.59 & \textbf{25.41} &                  29.82 &       36.27 \\
 CIFAR-10 &       DenseNet161 &         25.30 &        38.11 & \textbf{16.86} &                  27.21 &                  24.37 &                  26.92 &       34.27 \\
 CIFAR-10 &       DenseNet169 &         20.25 &        34.25 & \textbf{16.18} &                  22.95 &                  41.38 &                  36.91 &       26.38 \\
 CIFAR-10 &         GoogleNet &         19.85 &        24.06 &                  24.77 & \textbf{13.08} &                  16.88 &                  25.35 &       42.19 \\
 CIFAR-10 &       InceptionV3 &         31.42 &        80.60 & \textbf{17.31} &                  31.74 &                  20.06 &                  33.00 &       28.17 \\
 CIFAR-10 &       MobileNetV2 &         33.59 &        68.23 &                  25.32 & \textbf{24.74} &                  32.21 &                  74.59 &       27.27 \\
 CIFAR-10 &          ResNet18 &         34.79 &        28.21 & \textbf{17.80} &                  28.45 &                  76.02 &                  24.15 &       24.30 \\
\bottomrule
\end{tabular}%
}
\caption{MCE \% -- Maximum Calibration Error (lower is better), 15 bins, group 1}
\label{tab:metrics:MCE_1}
\end{table}

%% file: tabs/metrics_MCE_2.tex
\begin{table}[h!]
\centering
\resizebox{\textwidth}{!}{\begin{tabular}{llccccccc}
\toprule
Dataset &            Model &          Before calibration &           Hist-binning &               Isotonic &              T-scaling &              V-scaling &            V-scaling-b &            M-scaling-b \\
\midrule
    CIFAR-10 &            DenseNet121 (CE) &                  46.58 &        53.58 &                  23.62 & \textbf{22.26} &                  29.52 &                  23.21 &                  67.72 \\
    CIFAR-10 &          DenseNet121 (FL 1) &                  36.54 &        37.23 &                  26.98 &                  29.38 &                  31.21 &                  70.77 & \textbf{22.20} \\
    CIFAR-10 &          DenseNet121 (FL 2) &                  27.80 &        35.64 & \textbf{10.22} &                  66.85 &                  29.08 &                  28.61 &                  29.06 \\
    CIFAR-10 &          DenseNet121 (FL 3) &                  77.08 &        38.19 &                  38.07 &                  76.59 &                  74.49 & \textbf{19.89} &                  23.57 \\
    CIFAR-10 &       DenseNet121 (LS 0.05) &                  55.74 &        56.99 &                  80.80 &                  70.26 &                  57.18 &                  66.81 & \textbf{40.70} \\
    CIFAR-10 &              ResNet110 (CE) &                  47.15 &        50.61 &                  35.38 & \textbf{25.12} &                  25.90 &                  27.26 &                  31.14 \\
    CIFAR-10 &            ResNet110 (FL 1) &                  73.35 &        61.93 &                  26.03 &                  49.35 & \textbf{19.64} &                  42.96 &                  29.91 \\
\bottomrule
\end{tabular}%
}
\caption{MCE \% -- Maximum Calibration Error (lower is better), 15 bins, group 2}
\label{tab:metrics:MCE_2}
\end{table}

%% file: tabs/metrics_NLL_1.tex
\begin{table}[h!]
\centering
\resizebox{\textwidth}{!}{\begin{tabular}{llccccccc}
\toprule
Dataset &            Model &          Before calibration &           Hist-binning &               Isotonic &              T-scaling &              V-scaling &            V-scaling-b &            M-scaling-b \\
\midrule
 CIFAR-10 &       DenseNet121 &         0.257 &        0.427 &    0.354 &                  0.257 &                  0.258 & \textbf{0.255} &                  0.256 \\
 CIFAR-10 &       DenseNet161 &         0.261 &        0.495 &    0.285 &                  0.262 &                  0.255 & \textbf{0.253} &                  0.281 \\
 CIFAR-10 &       DenseNet169 &         0.245 &        0.559 &    0.309 &                  0.245 & \textbf{0.242} &                  0.242 &                  0.258 \\
 CIFAR-10 &         GoogleNet &         0.246 &        0.290 &    0.300 &                  0.241 &                  0.241 &                  0.239 & \textbf{0.235} \\
 CIFAR-10 &       InceptionV3 &         0.265 &        0.550 &    0.349 &                  0.266 &                  0.264 & \textbf{0.264} &                  0.264 \\
 CIFAR-10 &       MobileNetV2 &         0.241 &        0.494 &    0.309 &                  0.239 &                  0.239 & \textbf{0.237} &                  0.239 \\
 CIFAR-10 &          ResNet18 &         0.269 &        0.616 &    0.327 &                  0.269 &                  0.265 & \textbf{0.263} &                  0.269 \\
\bottomrule
\end{tabular}%
}
\caption{Negative Log-Likelihood (lower is better), group 1}
\label{tab:metrics:NLL_1}
\end{table}

%% file: tabs/metrics_NLL_2.tex
\begin{table}[h!]
\centering
\resizebox{\textwidth}{!}{\begin{tabular}{llccccccc}
\toprule
Dataset &            Model &          Before calibration &           Hist-binning &               Isotonic &              T-scaling &              V-scaling &            V-scaling-b &            M-scaling-b \\
\midrule
    CIFAR-10 &            DenseNet121 (CE) &         0.422 &        0.398 &    0.213 &                  0.216 &                  0.209 &                  0.206 & \textbf{0.198} \\
    CIFAR-10 &          DenseNet121 (FL 1) &         0.230 &        0.508 &    0.271 &                  0.190 &                  0.185 & \textbf{0.182} &                  0.193 \\
    CIFAR-10 &          DenseNet121 (FL 2) &         0.186 &        0.548 &    0.278 &                  0.181 &                  0.178 & \textbf{0.178} &                  0.182 \\
    CIFAR-10 &          DenseNet121 (FL 3) &         0.193 &        0.369 &    0.312 &                  0.194 &                  0.191 & \textbf{0.191} &                  0.206 \\
    CIFAR-10 &       DenseNet121 (LS 0.05) &         0.306 &        0.389 &    0.334 &                  0.306 &                  0.298 & \textbf{0.293} &                  0.302 \\
    CIFAR-10 &              ResNet110 (CE) &         0.512 &        0.454 &    0.227 &                  0.230 &                  0.224 & \textbf{0.222} &                  0.228 \\
    CIFAR-10 &            ResNet110 (FL 1) &         0.234 &        0.547 &    0.285 &                  0.181 &                  0.181 & \textbf{0.177} &                  0.182 \\
\bottomrule
\end{tabular}%
}
\caption{Negative Log-Likelihood (lower is better), group 2}
\label{tab:metrics:NLL_2}
\end{table}

%% file: tabs/metrics_BS_1.tex
\begin{table}[h!]
\centering
\resizebox{\textwidth}{!}{\begin{tabular}{llccccccc}
\toprule
  Dataset &            Model &          Before calibration &           Hist-binning &               Isotonic &              T-scaling &              V-scaling &            V-scaling-b &            M-scaling-b \\
\midrule
 CIFAR-10 &       DenseNet121 &                  0.102 &        0.106 & \textbf{0.099} &                  0.102 &     0.102 &                  0.102 &       0.101 \\
 CIFAR-10 &       DenseNet161 &                  0.102 &        0.107 & \textbf{0.098} &                  0.103 &     0.101 &                  0.101 &       0.104 \\
 CIFAR-10 &       DenseNet169 &                  0.098 &        0.106 & \textbf{0.095} &                  0.099 &     0.098 &                  0.098 &       0.100 \\
 CIFAR-10 &         GoogleNet & \textbf{0.109} &        0.112 &                  0.109 &                  0.109 &     0.110 &                  0.109 &       0.110 \\
 CIFAR-10 &       InceptionV3 &                  0.108 &        0.115 & \textbf{0.105} &                  0.109 &     0.109 &                  0.108 &       0.109 \\
 CIFAR-10 &       MobileNetV2 &                  0.102 &        0.106 & \textbf{0.099} &                  0.103 &     0.103 &                  0.102 &       0.103 \\
 CIFAR-10 &          ResNet18 &                  0.117 &        0.124 & \textbf{0.113} &                  0.116 &     0.115 &                  0.115 &       0.117 \\
\bottomrule
\end{tabular}%
}
\caption{Brier Score (lower is better), group 1}
\label{tab:metrics:BS_1}
\end{table}

%% file: tabs/metrics_BS_2.tex
\begin{table}[h!]
\centering
\resizebox{\textwidth}{!}{\begin{tabular}{llccccccc}
\toprule
Dataset &            Model &          Before calibration &           Hist-binning &               Isotonic &              T-scaling &              V-scaling &            V-scaling-b &            M-scaling-b \\
\midrule
    CIFAR-10 &            DenseNet121 (CE) &         0.094 &        0.094 & \textbf{0.081} &                  0.084 &                  0.082 &                  0.082 &                  0.082 \\
    CIFAR-10 &          DenseNet121 (FL 1) &         0.086 &        0.092 & \textbf{0.079} &                  0.081 &                  0.080 &                  0.080 &                  0.084 \\
    CIFAR-10 &          DenseNet121 (FL 2) &         0.081 &        0.089 & \textbf{0.079} &                  0.080 &                  0.079 &                  0.079 &                  0.080 \\
    CIFAR-10 &          DenseNet121 (FL 3) &         0.089 &        0.091 & \textbf{0.088} &                  0.089 &                  0.088 &                  0.088 &                  0.090 \\
    CIFAR-10 &       DenseNet121 (LS 0.05) &         0.102 &        0.102 & \textbf{0.091} &                  0.103 &                  0.102 &                  0.101 &                  0.100 \\
    CIFAR-10 &              ResNet110 (CE) &         0.097 &        0.098 &                  0.086 &                  0.087 &                  0.086 &                  0.086 & \textbf{0.086} \\
    CIFAR-10 &            ResNet110 (FL 1) &         0.086 &        0.088 &                  0.078 &                  0.079 &                  0.078 & \textbf{0.078} &                  0.078 \\
\bottomrule
\end{tabular}%
}
\caption{Brier Score (lower is better), group 2}
\label{tab:metrics:BS_2}
\end{table}